# Building and Displaying Name Relations using Automatic Unsupervised Analysis of Newspaper Articles

Bruno Pouliquen, Ralf Steinberger, Camelia Ignat, Tamara Oellinger

European Commission, Joint Research Centre
21020 Ispra (VA), Italy
http://www.jrc.it/langtech - http://press.jrc.it/NewsExplorer/ – Name.Surname@jrc.it

## Abstract

We present a tool that, from automatically recognised names, tries to infer inter-person relations in order to present associated people on maps. Based on an in-house Named Entity Recognition tool, applied on clusters of an average of 15,000 news articles per day, in 15 different languages, we build a knowledge base that allows extracting statistical co-occurrences of persons and visualising them on a per-person page or in various graphs.

## 1. Introduction

Being an international organisation, the European Commission needs to stay informed about news in Europe and in the world. Part of the information needed is about names, and more specifically about relations between names (person-person, organisation-person). The number of actors that appear in the news is increasing every day, and no one can manage to maintain a database of such "VIPs" unless it concerns a very specific area (NGOs, composition of governments, football teams, etc.). If we want to have a good coverage of relationships among names, we need to build an automatic information extraction tool from unstructured texts.

In the *Joint Research Centre* of the European Commission, the *Europe Media Monitor* (Best et al. 2002) gathers daily newspaper articles from different sources and different languages. Similar articles are grouped into clusters (a cluster could be seen as a story), and names are automatically extracted from each of these clusters. The result can be seen in the *NewsExplorer* tool, publicly available at http://press.jrc.it/NewsExplorer/.

The purpose of our tool, based on these names extracted from stories, is now to infer knowledge about relations between them using statistical measures. The advantage of such a tool is that it is automatically updated every day and that it can deal with a huge number of relations.

(Tufte 1998a) pointed out that the main reasons for displaying information in a graphical manner are the possibilities to show a big amount of data on a small space in order to maximise data density as well as make people understand large datasets. As over time a large amount of data has been collected through analysing millions of news articles, a suitable visualisation needed to be found. We have experimented with two different ways of displaying graphs of persons: a static image showing the most important relations regarding a given person, and an interactive graph, allowing the users to select the persons they want to look at.

Evaluating relationship relevance among people is extremely difficult and subjective; however we will try to measure the performance of our tool using a test set built out of the free Wikipedia online encyclopaedia (Wikipedia 2006).

## 2. Background and related work.

### 2.1 Relation maps

When a tool extracts person names from documents, it implicitly generates useful information regarding the co-occurrence of persons. Ben-Dov et al. (2004), who worked on both detecting relationships and visualising them, quote (Davies 1989): 'knowledge can be created by drawing inference from what is already known'. Such knowledge or information can be visualised with relation maps.

In principle, two methods can be used to automatically generate relation information: (a) the observation of the co-occurrence of names in the same text (at various levels: sentence, paragraph or text), and (b) the usage of syntactic-semantic rules to detect more specific relationships between persons. If two persons are often mentioned in the same document (co-occurrence information), they are likely to be in a certain relationship. This relationship is difficult to label, as it could be friendship, rivalry, family relationship, belonging to the same organisation, participation in the same meeting, etc. A rule-based system, on the other hand, would be able to detect more specific relationships. Ben-Dov et al. (2004) compare both approaches and come to the conclusion that, when searching for information about joint meetings, co-occurrence-based algorithms exhibit a good *Recall*, but are bad for *Precision*, while the inverse is true for rule-based methods. The authors estimate that writing rules to identify 'participation in a common meeting' takes a programmer between one and three weeks for one language only, assuming that an appropriate parser is available. The advantage of the co-occurrence-based approach - the method we use - is that no rules need to be written and that the same mathematical formulae can be used to describe relationships in any language.

To our knowledge, the closest project to ours is CORDER (Zhu et al. 2005) where the authors automatically extract entity links from texts taken from a web site. They tested it on web sites to highlight the persons that participate in different projects. Contrary to us, they used a manual evaluation, based on an existing ontology. In another technical report, Zhu et al. 2004 explain that the extraction of links is based on a clustering algorithm, which is relevant according to their evaluation. However, in our case we cannot apply such a heavy technique as the number of entities and documents is bigger than theirs (they had 2,164 entities and 1,863 documents; we have more than 85,000 names and 300,000 clusters to look at). Additionally we have to update our links every day. So we had to find a lighter method to be able to generate links for each person in less than a second.

The commercial system *Connivence Maps*, by *Pertinence Mining* (Connivence 2005) presents relationships among actors in the news, but they provide no details about the algorithms used.

The analysis of inter-concept links is part of the wider field of "link analysis". Examples of research in this field can be seen at the "Workshop on Link Analysis and Group Detection" (LinkKDD 2004).

*2.2 Gathering information from online Newspapers*

Our analysis is applied to the output of the *NewsExplorer* (Steinberger et al. 2005), part of the *Europe Media Monitor* system EMM (Best et al., 2002). EMM is a software toolset that monitors 800 different international news sources (in currently 30 languages) and extracts from them a daily average of 25,000 news articles. For a subset of about 15,000 articles per day in currently 13 languages (soon 15), we apply unsupervised hierarchical clustering techniques to group related articles separately for each language. We then track related news clusters within the same language and across seven of the languages. Our own Proper Name Recognition tool is applied to each of these clusters in order to extract names regarding the whole story (considering a cluster as a meta-text).

*2.3 Proper name recognition*

The proper name recognition has been described in Pouliquen et al. (2005). Here we give just a short description of this process. The names are recognised from plain text, using three methods (a) using a finite state automaton to recognised known names (b) using a list of common first names (e.g. 'John' followed by uppercased words) (c) using a semi-automatically built list of trigger words (e.g. 'president' followed by two uppercase words).

Our tool, being quite basic, has the advantage of being easily applied to any language using uppercase for proper names. Currently the tool works in English, French, Italian, Spanish, German, Portuguese, Swedish, Dutch, and to a certain extent also in Slovene, Estonian, Danish, Norwegian, Russian and Bulgarian. We also used the same tool but with more complex rules to detect names in Arabic (which does not distinguish uppercase and lowercase).

Additionally, we merge the various name variants of the same person. This aspect is very important when building relationship among people to avoid having two different entries for the same person.

*2.4 Storage of names in a database*

Finally, all names detected during our daily news analysis are stored in a database together with the cluster information. After one year of news analysis, the database has grown to about 230,000 distinct names (not counting variants of the same name), but only 85,000 of them are used more than once. On average, more than 500 new names are inserted every day. The trigger words found around the name are also stored. Variants of the same name are identified (see Pouliquen et al. 2005) and are stored with the same unique identifier.

*2.5 Trigger words*

To guess new names (method (b) described in Sect. 2.3), an extensive list of local patterns was developed in a boot-strapping procedure: We first wrote simple local patterns in PERL to recognise names in a collection of three months of English, French and German news. We then looked at the most frequent left and right hand side contexts of the resulting list of known names. For English alone, we currently have about 1,100 local patterns, consisting of titles ('Dr.', 'Mr', etc.), country adjectives (such as 'Estonian'), professions ('actor', 'tennis player', etc.), specific patterns (such as '[0-9]+ year-old'), etc. We refer to these local patterns as *trigger words*. For each added language, native speakers translate the existing pattern lists and use the same bootstrapping procedure to complete them.

Those patterns allow the program to recognise new names (i.e. in 'the American doctor John Smith'), but a stored list of such patterns is also useful to give users additional information about persons. In the previous example, for instance, the user will see that *John Smith* probably is an American doctor. When a name is often used with the same trigger words, statistical measures can be used to qualify names automatically. For instance, *George W. Bush* will be recognised as being the American president, *Rafik Hariri* as being the 'former Lebanese prime Minister', etc.

## 3. Getting information from names

The cluster and name information is stored in the NewsExplorer database. By grouping names by clusters we can generate additional semantically relevant information. The following sections show applications where names detected automatically from multilingual news collections are used.

### *3.1 Name browser*

In the JRC's NewsExplorer system, the information collected during the daily multilingual news analysis is stored in a relational database so that information about past events, persons and organisations can be browsed. For each cluster, in currently 13 languages (soon 15), the system keeps track which people are mentioned together with which other people, countries, and keywords. As the database is updated every day, a network of links builds up over time. For instance, the database can be queried for all news clusters that mention a certain person, and it can tell which other persons were mentioned in the same clusters. For each news cluster, a link to the original URL of the most typical article (the *medoid*, the one closest to the cluster centroid) allows the users to read up on the story.

A web interface gives access to the information stored about each person. This information includes:

- information about the person itself: name, name variants, photograph (when available);
- clusters this person was mentioned in;
- the trigger words (*titles*) most frequently identified in the context of this person;
- a list of *related* and *associated* persons, i.e. those persons that are frequently mentioned in the same news clusters, and the persons that are more specifically associated with him/her.

Additionally, a daily *VIP list* displays the persons most often mentioned in the news of that day.

As the titles are stored in the database, the user can also query all persons having a given *title* (e.g. 'Georgian president'). For details on the browsing functionalities, see Steinberger et al. (2005).

Most of the information is exported to a public web site (http://press.jrc.it/NewsExplorer/), where each name is displayed on its own webpage, as shown in Figure 1.

*Figure 1:* NewsExplorer entry showing the various variants of a name, the stories he/she appears in, the related and associated persons, and the titles found in the context of his/her name.

## *3.2 Identifying links between persons*

The basic way to compute relationship consists in computing the co-occurrence frequency of two names in the same clusters; we call the resulting list *related* people. The co-occurrence is currently computed at the cluster level. This means that, in theory, two persons could be associated even if they do not occur in the same newspaper article, but if their corresponding articles are part of the same cluster. Another option would have been to compute it at the document level, or even at the paragraph or sentence level.

When displaying the *related* persons (ranked by frequency of co-occurrence at the cluster level), the people that are frequently in the news (e.g. *George W. Bush*) will appear in almost all the lists. We therefore introduced a weighting factor that allows to down-weight highly frequent names and to focus on those person names that are specifically *associated* with a given person. We call this list the *associated* persons. The weighting formula uses three factors: the number of clusters each person appears in, the number of common clusters two persons appear in, and the number of 'further associates' each of the persons have. The formula computes a *specific association weight* between two entities in our database:

$$w_{e_1,e_2} = Co_{e_1,e_2} \cdot Icf_{e_1,e_2} \cdot Iass_{e_1,e_2}$$

**Equation 1**. *Relationship weight between two entities*

Where:

$e_i$: Entity *i*

$Co_{e1,e2}$: Cluster co-occurrence weight between $e_1$ and $e_2$

$Icf_{e1,e2}$: Inverse cluster frequency of $e_1$ and $e_2$

$Iass_{e1,e2}$: Inverse association frequency of $e_1$ and $e_2$

$$Co_{e_1,e_2} = 1 + \ln(C_{e_1,e_2})$$

**Equation 2**. *Cluster co-occurrence weight*

Where:

$C_{e1,e2}$: Number of clusters where $e_1$ and $e_2$ occur together

*Instead of choosing the frequency of co-occurrence, where two persons occurring 100 times together has a value twice as much as two persons occurring 50 times together, we prefer using this logarithmic scale (also called sometimes the TFlog value)*

$$Icf_{e_1,e_2} = \frac{2C_{e_1,e_2}}{\left(C_{e_1} + C_{e_2}\right)}$$

**Equation 3**. *Inverse cluster frequency*

Where:

$C_{e1,e2}$: Number of clusters where $e_1$ and $e_2$ appear together

$C_{ei}$: Total number of clusters where $e_i$ appears; i=1,2

$$Iass_{e_1,e_2} = \frac{1}{1 + \ln(A_{e_1}.A_{e_2})}$$

**Equation 4**. *Inverse association frequency*

Where:

$A_{ei}$: Total number of entities occurring with $e_i$; i=1,2

The *associated* person list (using weighted formula) shows rather different names from the *related* person list (based on co-occurrence frequency). For the Secretary-General of the Council of the European Union *Javier Solana*, for instance, the most frequently co-occurring names are the well-known politicians *George W. Bush*, *Jacques Chirac*, *Yasser Arafat* and *Kofi Annan*. In the weighted list, however, the two top-ranking persons are *Christina Gallach* (Solana's spokesperson) and *Pierre de Boissieu* (Solana's assistant). These two persons are less widely known because they are not mentioned much outside the context of Javier Solana, but their names are very closely linked to Solana as they are typically mentioned in the news when Solana is mentioned.

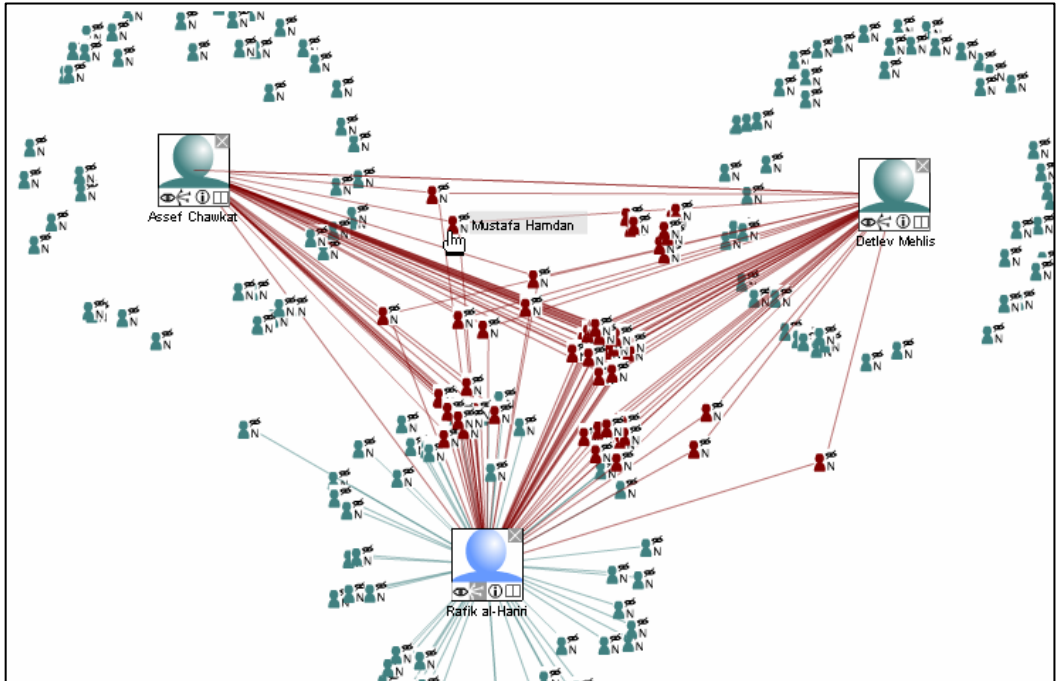

*Figure 2:* *NewsExplorer interactive map, showing three persons (Detlev Mehlis, judge appointed by the UN, for the Rafik Hariri assassination, showing implication of Syria's military intelligence chief Assef Chawkat) and the links via other persons (here Mustafa Hamdan, suspect in the murder case). Here the user, starting from the Rafik Hariri entity, has chosen to display also the entities Detlev Mehlis and Assef Chawkat entities, all three having a link with Mustafa Hamdan.*

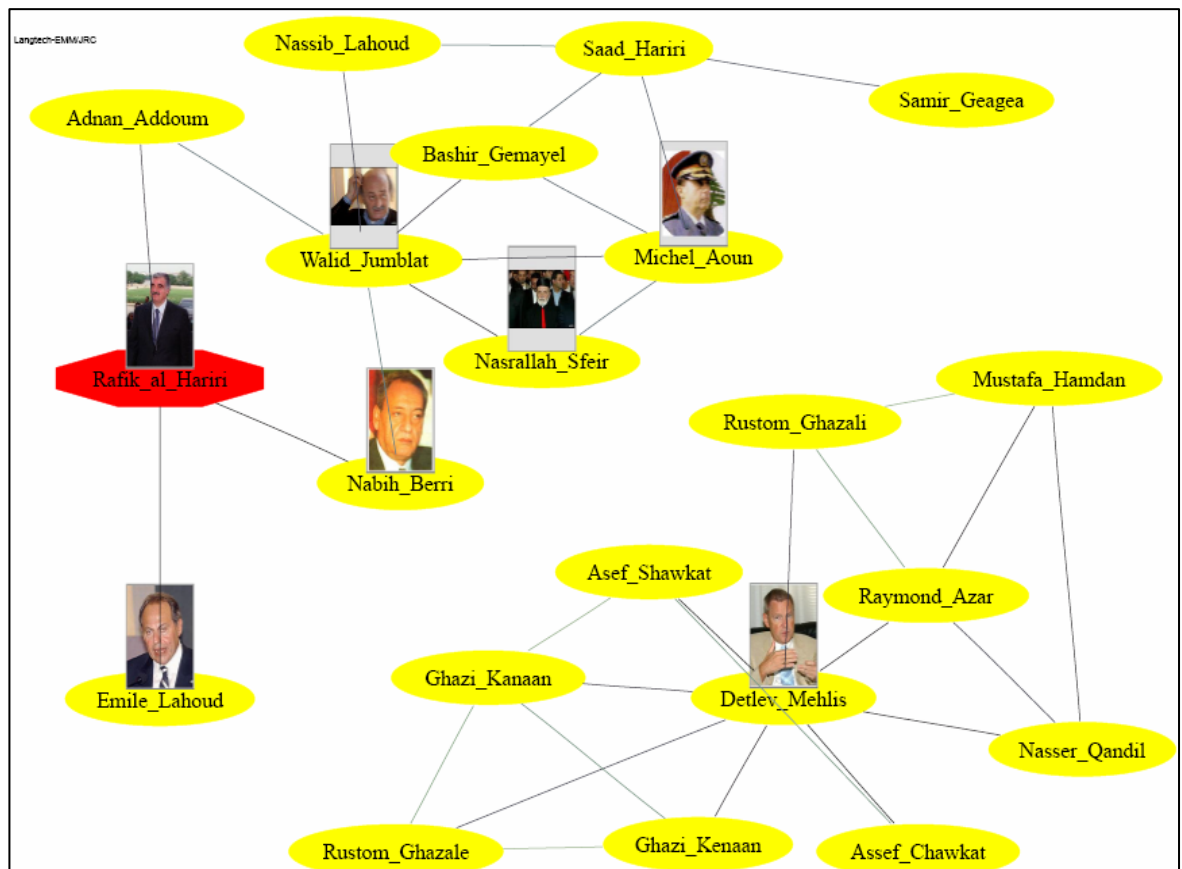

*Figure 3:* *Relation map showing Rafik Hariri and the first 18 persons associated to him (N=18)*

Another example can be seen in Figure 1 where, for Rafik Hariri, *George W Bush* and *Kofi Annan* appear in the list of '*related*' persons, but not in the '*associated*' list. We should mention that, even if the associating value is often less than 1.0, it has no upper limit.

### *3.3 Displaying relation maps*

#### *3.3.1 Using Flash interactive maps*

In the visualisation interface, each person and its relations is drawn as a *small multiple*[i], a graphic design that according to [Tufte 1998b] "visually enforces comparisons of changes, of the differences among objects, of the scope of alternatives". So relations of a single person can be compared to another person's relations. Furthermore the users are supported in their reasoning through the interface by automatically finding overlaps and highlighting those persons that are related to more than one person.

The interactive aspect of the interface allows the users to focus on the information that they are interested in. Certain relationships can be hidden, for each related person; their relationships can be expanded; additional information about a person can be seen by going directly from the graph to the person's page within NewsExplorer or the Wikipedia website. A search function allows looking for a certain name within the collected data. If the person is already in the graph, it will be highlighted, otherwise it can be added to the graph. As soon as a non-related person is added to the graph, relations in common will be found and highlighted. This interactive feature helps finding not just direct relations but also indirect relations via another person. The users are able to generate a visualisation according to their needs.

The visualisation interface was programmed using Macromedia Flash. This software allows producing high quality vector graphics with relatively small file sizes. The data is kept in XML format and as soon as a person is added to the image or the relations of a person are shown, the respective data is loaded into the application.

#### *3.3.2 Using static maps*

Another approach consists of showing a number of associations directly on a graph. We use the *graphviz* open source graph visualisation software[ii], and more specifically the *neato* utility based on the algorithm of [Kamada & Kawai, 1989], which uses a 2-D graph to display the closest nodes together. For a given person *A*, we give as input a non-oriented graph where each node is a person and each edge is a co-occurrence relationship. Our module takes as parameter a person and computes the undirected graph. A filter then allows displaying only the first *N* relations (those *N* relations having the highest weight). Figure 3 gives an overview of the persons associated with *Rafik Hariri*. The users can then select another person and display his or her corresponding graph. This graph is useful to give a quick overview of various groups of people related to the person.

---

[i] Small Multiples are arrangements of graphics with the same design but different datasets. It can be compared to the single frames of a movie.
[ii] http://www.graphviz.org/

## 4. Evaluation

The evaluation of the relevance of association between people is extremely subjective, even when done by human: Should we associate football players with other members of the team, or with children and wives, or with other organisations they belong to, etc…

Anyway we wanted to have an evaluation set as big as possible and as objective as possible. The names in our database are mentioned in the news, so hopefully well-known persons. In order to evaluate our system we decided to use an up-to-date encyclopaedia that contains people descriptions with links to related persons. To our knowledge the only big encyclopaedia freely available, and easily browsable, is Wikipedia (Wikipedia 2006).

### *4.1 Building confirmed relations among people from Wikipedia*

We built an automatic agent that looks for persons in our database and downloads the corresponding pages from Wikipedia (currently in English, French, German, Italian and Spanish). When available, the page is saved in the original HTML format; then a parser looks for hyperlinks and saves them in a second Xml file.

All the hyperlinks are analysed in order to know if they refer to a person or something else (Wikipedia hyperlinks are not qualified: we do not know if the hyperlink refers to a person entry, a date, a country, an image, a category, etc... We considered our name database to be exhaustive enough to cover most of the known persons, so we decided to stick to the hyperlinks referring to a string which is already known as being a person name.

At this stage we are able to generate all hyperlinks from a person page to another one. In order to confirm this hyperlink as really being a strong person-to-person relationship, we required to have the inverse link (i.e. if person B mentioned in person A's page, we need person A to be mentioned in B's page).

As a result of our process, we generated an Xml file containing 8206 relationship. This first set will be kept as a baseline for our evaluation. It contains 1109 different persons with an average of 5.51 relations per person (minimum 1, maximum 54 – for *John Kerry*).

We wanted to avoid "noise" in the baseline, moreover our application aims at being multilingual. We thus built a multilingual evaluation set where each inter-person relation has to be confirmed in two different languages in Wikipedia. Applying this additional criterion yields a subset of 1468 relations (only those available in at least two different languages). It contains 381 different persons with an average of 3.85 relations per person (minimum 1, maximum 27 – for *George W. Bush*).

Currently NewsExplorer's Named Entity Recognition tool is not optimised for organisations. We have thus chosen to exclude the relations to organisation names from our evaluation.

### *4.2 Comparing baseline with our automatically generated relations*

In order to evaluate the relevance of the *related* and *associated* persons, we launched comparisons with the two Wikipedia baselines. Each time we present the results in different columns for the *related* persons using co-occurrence frequency and for the *associated* persons (weighted formula). Each evaluation consisted in comparing the first *N* relations (*N* being called the rank) to the Wikipedia baseline. For each *N* rank, we compute the precision (ratio of automatic relations which are part of Wikipedia baseline), and the recall value (ratio of Wikipedia baseline relations among the automatic ones).

We show the results in Table 1 for the Wikipedia baseline of 8206 relations, and in Table 2 for the subset of 1468 *stronger* relations.

**Table 1: Evaluation of the comparison between *related/associated* persons to the 1468 relations of the Wikipedia 1468-baseline. As the baseline contains semantically strong relations (confirmed in two Wikipedia pages in different languages, the recall is quite high when considering enough relations.**

| Rank | *Related* persons (co-occurrence frequency) | | *Associated* persons (weight formula) | |
|---|---|---|---|---|
| | Precision | Recall | Precision | Recall |
| 01 | 0.01 | 0.01 | 0.02 | 0.01 |
| 02 | 0.02 | 0.05 | 0.04 | 0.02 |
| 03 | 0.03 | 0.08 | 0.04 | 0.04 |
| 04 | 0.04 | 0.11 | 0.04 | 0.05 |
| 05 | 0.04 | 0.14 | 0.04 | 0.05 |
| 10 | 0.04 | 0.23 | 0.04 | 0.10 |
| 20 | 0.04 | 0.38 | 0.03 | 0.18 |
| 30 | 0.04 | 0.52 | 0.03 | 0.27 |
| 50 | 0.03 | **0.69** | 0.03 | 0.37 |
| 75 | 0.03 | 0.79 | 0.02 | 0.46 |
| 100 | 0.02 | 0.80 | 0.02 | 0.54 |

**Table 2: Evaluation of the comparison between *related/associated* persons to the "big" Wikipedia 8206-baseline. The precision is higher than the 1468 baseline (because the number of relations is bigger), but the recall is lower due to the weakness of the relations when relying on one language only in the Wikipedia pages.**

| Rank | *Related* persons (co-occurrence Frequency) | | *Associated* persons (weight formula) | |
|---|---|---|---|---|
| | Precision | Recall | Precision | Recall |
| 01 | **0.05** | 0.02 | **0.09** | 0.01 |
| 02 | 0.06 | 0.05 | 0.08 | 0.03 |
| 03 | 0.07 | 0.07 | 0.07 | 0.04 |
| 04 | 0.06 | 0.10 | 0.06 | 0.05 |
| 05 | 0.07 | 0.12 | 0.06 | 0.06 |
| 10 | 0.06 | 0.19 | 0.05 | 0.10 |
| 20 | 0.05 | 0.32 | 0.04 | 0.15 |
| 30 | 0.05 | 0.41 | 0.03 | 0.21 |
| 50 | 0.04 | 0.54 | 0.03 | 0.29 |
| 75 | 0.03 | 0.63 | 0.02 | 0.36 |
| 100 | 0.03 | 0.64 | 0.02 | 0.43 |

The evaluation shows that our system can find the expected related persons (in about 70% of the cases when looking on the first 50 associated persons - recall at 0.69 for rank 50). We were not expecting such a good result knowing that the content of our source is often very detailed, especially when listing relations that are not mentioned any more in the news (e.g. George W. Bush's Wikipedia page contains reference to the 42 previous US presidents).

When comparing to this baseline, the system is still not very precise (precision at maximum 0.09). Of course we have to balance this result by the fact that almost all the persons associated with the weighted formula have indeed clear relationship with the first entity, only that, of course, Wikipedia cannot list all of them (the Wikipedia page for European Commission president José Manuel Durao Barroso does not contain direct reference to any of the European Commissioners, and most often the commissioners' page does not have direct reference to Barroso). Furthermore, clearly related persons, like Solana's secretary and spokesperson, are not mentioned in Solana's Wikipedia page.

An outcome of the evaluation is that the weighted formula produces better precision values than the co-occurrence frequency formula for the lowest ranks. For rank 1 it is even twice as high (0.09 instead of 0.05 for the 8206-baseline, 0.02 instead of 0.01 for the 1468-baseline).

## 5. Conclusion and future work

This work allows the JRC to generate links between names using automatically gathered unstructured texts. The generation of such information is now completely automatic and updated every morning. This information is now accessible online since August 2005 as part of the *NewsExplorer*. Each recognised name has now its own web page, showing the related and associated names, among other information. Additionally to this person's webpage, the interactive visualisation interface allows exploring the direct relations as well as indirect relations to other persons. Users can "navigate" through a large dataset. Starting with a first overview they can go more and more into detail, exploring the information bit by bit. So the users are not only recipients, they interactively design their own network visualisation with focus on interesting features.

The computed relationships have been shown to be relevant using an external knowledge source, the Wikipedia encyclopaedia. This standard is highly questionable by itself, so an additional manual evaluation should be done in order to judge the relevance of the associations which are not part of Wikipedia entries.

The evaluation also shows that our weighted association formula has not a satisfying recall. Varying the parameters and launching an evaluation on our test set could raise better results. Additionally, the association of persons could benefit from the sentence-level co-occurrence (mentioning two persons in the same sentence is more significant than in the same story).

Our tool has been put online recently, we now have to identify users' requirements in order to target the next developments, some areas of interest could be social network discovery, highlighting political relationship among peoples…

A major challenge now is to label the relationships between people. We currently know that two persons are linked, but we are not able to label the type of this link (family relation, part of the same organisation, etc.). It would be a major improvement when displaying the graphs. We are thinking, as a minimal feature, to allow the user to browse the sentences where the two names appear together.

Another way of improving the precision of our tool would be to use the Wikipedia links as part of our system. The relationship between people, for instance, could be confirmed if a given person is mentioned in somebody else's page.

## Acknowledgements

We thank the whole team of the JRC Web Technology sector for providing us with the valuable news data to test the tools, as well as for their technical support. Without forgetting the numerous persons who helped us with their native speaker knowledge.

## References

Ben-Dov M., Wu W., Feldman R. & Cairns P. (2004), *Improving knowledge discovery by combining text-mining and link analysis techniques*. SIAM International. Conference. on Data Mining, Florida, USA.

Best C., van der Goot E., de Paola M., Garcia T. & Horby D. (2002). *Europe Media Monitor – EMM*. JRC Technical Note No. I.02.88. http://emm.jrc.cec.eu.int

Connivence (2005) *Connivence maps: relationships between actors from news feeds*. Available at http://www.connivences.info/, (last visited 06/06/2005).

Kamada T. and Kawai S. (1989) *An algorithm for drawing general undirected graphs*. Information Processing Letters, 31(1):7–15, April 1989.

Link KDD workshop (2004), *Workshop on Link Analysis and Group Detection*, part of KDD 2004 conference, Seattle, WA, USA, 22 August 2004, available at http://www.cs.cmu.edu/~dunja/LinkKDD2004/ (last visited 14/11/2005)

Pouliquen B., Steinberger R., Ignat C., Temnikova I., Widiger A., Zaghouani W. & Žižka J. (2005). *Multilingual person name recognition and transliteration*. Revue CORELA, Cognition, Représentation, Langage.

Steinberger R., Pouliquen B. & Ignat C. (2005). *Navigating multilingual news collections using automatically extracted information*. Journal of Computing and Information technology. CIT 13, 2005, 4, pp. 257-264, available at http://cit.zesoi.fer.hr/downloadPaper.php?paper=767

Tufte E. R. (1998a) *The Visual Display of Quantitative Information*. Graphics Press, Cheshire, Connecticut, 16[th] edition 1998

Tufte E. R. (1998b) *Envisioning Information*. Graphics Press, Cheshire, Connecticut, 6[th] edition, 1998

Wikipedia (2006), *Multilingual Web-based free-content encyclopedia*, http://www.wikipedia.org/ (last visited 23/01/2006)

Zhu Jianhan, Alexandre Gonçalves, Victoria Uren (2004) *Adaptive Named Entity Recognition for Social Network Analysis and Domain Ontology Maintenance*, Technical Report kmi-04-30. December 2004. available at http://kmi.open.ac.uk/publications/pdf/kmi-04-30.pdf (last visited 14/11/2005)

Zhu J., Gonçalves A., Uren V., Motta E., Pacheco R. (2005): *CORDER: COmmunity Relation Discovery by named Entity Recognition*, Proc. of Third International Conference on Knowledge Capture (K-Cap'2005), October 2-5, 2005 Banff, Canada